%% file: main.tex
\documentclass[]{template}

\usepackage[utf8]{inputenc}             % allow utf-8 input
\usepackage[T1]{fontenc}                % use 8-bit T1 fonts
\usepackage{url}                        % simple URL typesetting
\usepackage{booktabs}                   % professional-quality tables
\usepackage{multirow}
\usepackage{colortbl}
\usepackage{multicol}
\usepackage{amsfonts}                   % blackboard math symbols
\usepackage{nicefrac}                   % compact symbols for 1/2, etc.
\usepackage{microtype}                  % microtypography
\usepackage[dvipsnames]{xcolor}         % colors

\usepackage{latexsym}

\usepackage{graphicx}
\usepackage{float}
\usepackage{subcaption}
\usepackage{wrapfig}
\usepackage{lipsum}
\usepackage{adjustbox}

\usepackage{bm}

\usepackage{tabularx} 
\usepackage{ragged2e} 
\newcolumntype{L}{>{\RaggedRight\hangafter=1\hangindent=0em}X}

\usepackage{enumitem}

\usepackage{amsmath}
\usepackage{amssymb}
\usepackage{mathtools}
\usepackage{amsthm}

\setboolean{logo}{true}    % 设为 true 表示显示 logo；设为 false 则不显示

\usepackage[linesnumbered,ruled,vlined]{algorithm2e}

\hypersetup{
    colorlinks=true,
    linkcolor=red,
    citecolor=Cerulean,
    filecolor=magenta,      
    urlcolor=magenta,
}

\usepackage[capitalize,noabbrev]{cleveref}
\crefname{section}{§}{§§}
\Crefname{section}{§}{§§}

\usepackage{calligra}
\DeclareMathAlphabet{\mathcalligra}{T1}{calligra}{m}{n}

\usepackage{pifont}

\theoremstyle{plain}

\theoremstyle{definition}

\theoremstyle{remark}

\renewcommand{\paragraph}[1]{\vspace{1mm}\noindent\textbf{#1}}

\DeclareCaptionLabelFormat{cont}{#1~#2\alph{ContinuedFloat}}
\usepackage[most]{tcolorbox}
\tcbset{
  promptbox/.style={
    top=10pt,
    colback=lightgray!20,
    colframe=Black,
    colbacktitle=NavyBlue,
    enhanced,
    center,
    attach boxed title to top center={yshift=-0.1in,xshift=0.0in},
    boxed title style={boxrule=0pt,colframe=white,},
  }
}
\newtcolorbox{promptbox}[2][]{promptbox, title=#2,#1}
\tcbset{
  takeawaybox/.style={
    top=10pt,
    colback=lightgray!20,
    colframe=Black,
    colbacktitle=BurntOrange,
    enhanced,
    center,
    attach boxed title to top center={yshift=-0.1in,xshift=0.0in},
    boxed title style={boxrule=0pt,colframe=white,},
  }
}
\newtcolorbox{takeawaybox}[2][]{takeawaybox, title=#2,#1}
\tcbset{
  observationbox/.style={
    top=10pt,
    colback=lightgray!20,
    colframe=Black,
    colbacktitle=YellowGreen,
    enhanced,
    center,
    attach boxed title to top center={yshift=-0.1in,xshift=0.0in},
    boxed title style={boxrule=0pt,colframe=white,},
  }
}
\newtcolorbox{observationbox}[2][]{observationbox, title=#2,#1}

\usepackage{xspace}

\newcommand\blfootnote[1]{%
  \begingroup
  \renewcommand\thefootnote{}\footnote{#1}%
  \addtocounter{footnote}{-1}%
  \endgroup
}

\usepackage{CJK}

\title{Intern-S1-Pro: Scientific Multimodal Foundation Model at Trillion Scale}

\author[]{Intern-S1-Pro Team, Shanghai AI Laboratory}

\input{sections/0.abstract}

\begin{document}

% \blfootnote{$\dagger$ Corresponding authors: AAA (AAA@pjlab.org.cn), BBB (BBB@pjlab.org.cn)}
\blfootnote{$*$ Model is available at \url{https://huggingface.co/internlm/Intern-S1-Pro}}

\maketitle

\input{sections/1.introduction}
\input{sections/2.arch}
\input{sections/3.pretrain}
\input{sections/4.posttrain}
\input{sections/5.evaluation}

\input{sections/6.conclusion}

\clearpage
\bibliographystyle{plain}
\bibliography{refs}

%%%%%%%%%%%%%%%%%%%%%%%%%%%%%%%%%%%%%%%%%%%%%%%%%%%%%%%%%%%%

\clearpage
%\appendix
%\input{sections/appendix}

%%%%%%%%%%%%%%%%%%%%%%%%%%%%%%%%%%%%%%%%%%%%%%%%%%%%%%%%%%%%

% \newpage
% \input{sections/checklists}

\end{document}

%% file: sections/0.abstract.tex
\begin{abstract}

We introduce Intern-S1-Pro, the first one-trillion-parameter scientific multimodal foundation model. Scaling to this unprecedented size, the model delivers a comprehensive enhancement across both general and scientific domains. Beyond stronger reasoning and image-text understanding capabilities, its intelligence is augmented with advanced agent capabilities. Simultaneously, its scientific expertise has been vastly expanded to master over 100 specialized tasks across critical science fields, including chemistry, materials, life sciences, and earth sciences. Achieving this massive scale is made possible by the robust infrastructure support of XTuner and LMDeploy, which facilitates highly efficient Reinforcement Learning (RL) training at the 1-trillion parameter level while ensuring strict precision consistency between training and inference. By seamlessly integrating these advancements, Intern-S1-Pro further fortifies the fusion of general and specialized intelligence, working as a Specializable Generalist, demonstrating its position in the top tier of open-source models for general capabilities, while outperforming proprietary models in the depth of specialized scientific tasks. %Especially, Intern-S1-Pro outperforms the specialist model that is designed for specific scenarios in several scientific tasks. 

\end{abstract}

%% file: sections/1.introduction.tex
\section{Introduction}

The advent of Large Language Models (LLMs) and Visual Language Models (VLMs) has fundamentally transformed the landscape of artificial intelligence, offering unprecedented capabilities in reasoning, generation, and multimodal understanding~\cite{achiam2023gpt, kaplan2020scaling}. In the domain of AI for Science (AI4S), these foundation models have emerged as critical tools for accelerating scientific discovery, enabling researchers to tackle complex problems ranging from protein structure prediction to materials design~\cite{zhang2023scientific, taylor2022galactica, merchant2023scaling}. Large Models serve as a unified interface for processing vast amounts of scientific literature, experimental data, and domain-specific knowledge, thereby bridging the gap between disparate scientific disciplines~\cite{singhal2023large}.

To build an effective scientific foundation model, scaling model size is imperative due to the immense diversity inherent in scientific domains. Compared to natural language, science encompasses much more specialized fields, such as chemistry, biology, physics, and earth sciences, each with its own unique ``language", including domain-specific notations, knowledge, and reasoning patterns. This diversity grows with diving into the frontier science since they often involve long-tailed knowledge and specialized skills. Previous works in multilingual machine translation has founded that a single model requires more parameters when asking it to translate more language pairs, such as the model size for hundreds of language pairs is 90x larger compared to a bilingual model~\cite{nllb2022}. A scientific foundation model should possess sufficient capacity to master a wide array of scientific tasks while retaining general text and vision capabilities. 

In this work, we introduce \textbf{Intern-S1-Pro}, the first one-trillion-parameter scientific multimodal foundation model. Scaling to this unprecedented size, Intern-S1-Pro delivers a comprehensive enhancement across both general and scientific domains. Beyond stronger reasoning and image-text understanding capabilities, its intelligence is augmented with advanced agent capabilities, enabling it to autonomously plan and execute complex scientific workflows. Simultaneously, its scientific expertise has been vastly expanded to master over 100 specialized tasks across critical science fields, including chemistry, materials, life sciences, and earth sciences. 

\begin{figure}[ht]
    \centering
    \includegraphics[width=0.85\linewidth]{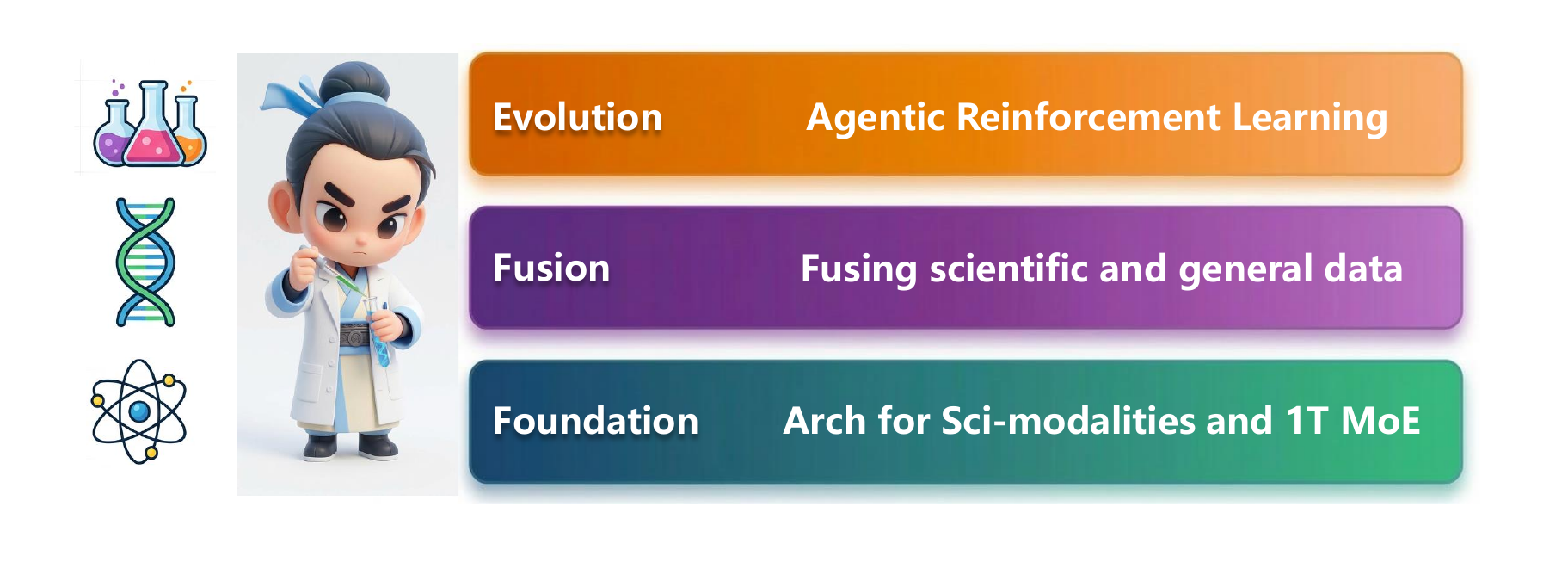}
    \caption{The SAGE (Synergistic Architecture for Generalizable Experts, including three layers, Foundation, Fusion, and Evolution) framework used in Intern-S1-Pro development, illustrating the core capabilities and the integrated learning process that enables synergistic improvements across domains.}
    \label{fig:intro_sage}
\end{figure}

Following the three layers design in SAGE framework (shown in Figure~\ref{fig:intro_sage}), we demonstrate that Intern-S1-Pro, through joint training on general and specific tasks, can outperform specialized models in several scientific tasks. Contrary to the common belief that specialized models are superior for niche tasks, our findings reveal that a sufficiently large generalist model, when trained jointly, can achieve superior performance. In Section~\ref{sec:bio_case}, we will show in detail that even when using similar training data to specialized models, utilizing a larger model architecture and a joint training strategy yields significant performance gains, validating the effectiveness of our approach.

Scaling model parameters introduces new challenges, and we list two architecture-related problems here: training instability due to load imbalance among a massive number of experts, and the difficulty in sufficiently optimizing router embeddings. Extreme imbalance of experts can cause memory spikes, and the conventional solution is to adopt a robust but slower parallelism strategy. To preserve both the stability and the efficiency, we propose a \textbf{group routing mechanism} that enforces a lower bound on expert load balance. Additionally, while we initialize experts from Intern-S1 to ensure a strong starting point of experts' (the Fully Forward Network part) weights, the router embeddings require efficient learning to handle the expanded expert pool; thus, we introduce a \textbf{gradient estimation scheme} to accelerate their update frequency.

On the engineering front, maintaining high training throughput is critical. Through the co-design of algorithms and infrastructure, we achieve deep optimization between the XTuner training framework and the LMDeploy inference engine. This synergy allows Intern-S1-Pro to scale to $4\times$ the size of its predecessor (Intern-S1) while incurring only a $\sim20\%$ reduction in training efficiency. This demonstrates that with careful system-level optimizations, massive-scale training can remain highly efficient. Meanwhile, the robust and optimized infrastructure facilitates highly efficient RL training at the 1-trillion parameter level while ensuring strict precision consistency between training and inference.

%% file: sections/2.arch.tex
\section{Architecture}

\begin{figure}
    \centering
    \includegraphics[width=0.9\textwidth]{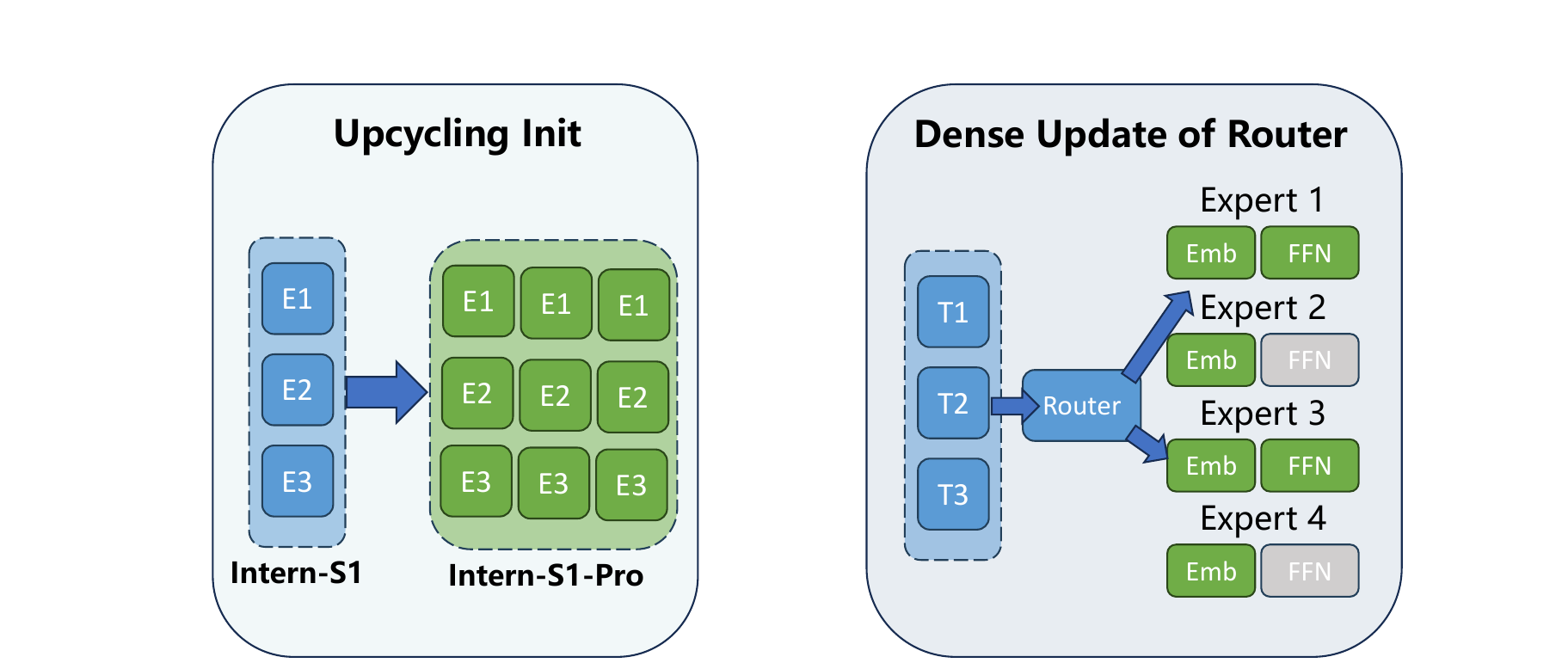}
    \caption{Left: Illustration of the expert expansion process from Intern-S1 to Intern-S1-Pro. The grouped routing strategy ensures well-trained Top-1/Top-2 experts are distributed across groups to maintain training stability. Right: The illustration of Straight-Through Estimator for enabling updating all router embeddings in every pass.}
    \label{fig:init_dense_grad}
\end{figure}

Intern-S1-Pro is derived from Intern-S1 through expert expansion, as illustrated in Figure \ref{fig:init_dense_grad}. In this expansion process, we incorporate the Grouped Routing design, where experts are distributed into groups. We ensure that the experts activated within each group correspond to the Top-1 or Top-2 experts prior to expansion. While this approach results in some homogenization of expert activation during the initialization phase, the experts naturally differentiate after a few step training, and this design significantly enhances training stability. In contrast, assigning differentiated experts corresponding to the pre-expansion Top-1 to Top-8 across groups leads to training instability and performance degradation. For example, we tested two initialization methods on a 30BA3 model and 2000 training steps. The first method can slight outperform the model prior to expansion while the second method have a performance drop over 20pts. Our hypothesis is that the experts that often activated as top-1 selection suggest they are well-trained and important modules, so keeping each group has well-trained experts are essential to the initialization.

\subsection{Group Routing}

\begin{figure}[t]
    \centering
    \includegraphics[width=0.6\textwidth]{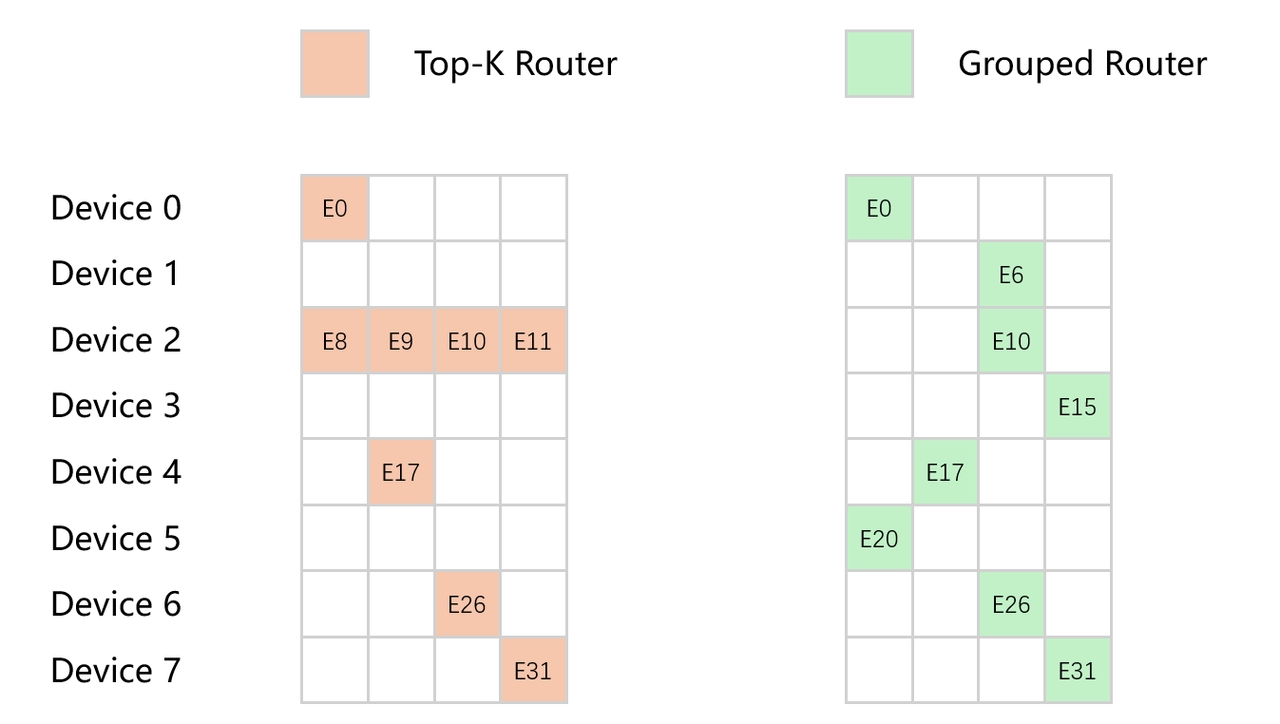}
    \caption{Training with Grouped Router can achieve absolute load balancing across devices for MoE models with a Top-k configuration of k=8 under the EP8 training strategy.}
    \label{fig:grouped_router}
\end{figure}

For the training of ultra-large-scale MoE models (e.g., Intern-S1-Pro), Expert Parallelism (EP) serves as the core technical approach to mitigate GPU memory and communication overheads. However, the expert load imbalance caused by the traditional Top-$k$ routing strategy will lead to cross-device load imbalance during the expert parallel training process. Although a lower degree of expert parallelism (e.g., EP8) and the MoE Balance Loss can alleviate this issue, the phenomenon still persists, and is particularly severe in the post-training phase of large models. This not only significantly degrades the training efficiency of expert parallelism but also causes an Out-of-Memory (OOM) risk during training in extreme cases.

To address the aforementioned problem, we propose to replace the traditional Top-K Router with the Grouped Router to achieve absolute load balancing across devices under the 8-way expert parallelism training strategy, thereby stabilizing the training process and improving training efficiency. Specifically, let the total number of experts in an MoE layer be denoted as $E$, and the expert parallelism degree  be $S$. In Grouped Router architecture, all experts are uniformly partitioned into $G$ mutually disjoint groups based on device mapping, denoted as $\{\mathcal{E}_1, \mathcal{E}_2, \ldots, \mathcal{E}_G\} $, with each group containing $E/G$ experts. For each group $g$, only the top-($K/G$) experts with the highest scores are selected within the group, and the final set of activated experts is obtained by taking the union of these intra-group top experts, as illustrated in Figure \ref{fig:grouped_router}. Combined with the configuration of the Intern s1-pro 1T model ($k=8$) and the EP8 training strategy, we can divide all experts into 8 groups and select the Top-1 expert within each group, ultimately achieving absolute load balancing across devices. This approach not only significantly improves training efficiency but also fundamentally eliminates the OOM risk during training.

\subsection{Straight-Through Estimator for Sparse Expert Routing}

MoE architectures scale model capacity by routing each input token to a small subset of $K$ out of $N$ experts via Top-$K$ selection. Given the token representation $\mathbf{x}\in\mathbb{R}^{d}$ and the router parameter $\mathbf{W}_r\in\mathbb{R}^{N\times d}$, the router produces logits $\mathbf{z}=\mathbf{W}_r\mathbf{x}$, computes routing probabilities $\mathbf{p}=\mathrm{softmax}(\mathbf{z})$, and selects $S=\mathrm{TopK}(\mathbf{p},K)$. The layer output is:
\begin{equation}
\mathbf{y}=\sum_{i\in S}\tilde{p}_i \cdot E_i(\mathbf{x}),
\label{eq:moe_output}
\end{equation}
where $\tilde{p}_i = p_i / \sum_{j\in S} p_j$ is the normalized routing weight for the $i$-th expert network $E_i$.

We introduce the Straight-Through Estimator (STE)~\cite{hinton2012neural,bengio2013estimating} to decouple the forward and backward passes of the routing operation~\cite{yao2025densemixer,liu2024grin,liu2023sparse}.
In the forward pass, the standard sparse Top-$K$ selection is preserved exactly.
In the backward pass, gradients flow through the softmax distribution without re-normalization.
The STE routing weight is constructed as:
\begin{equation}
\hat{p}_i^{\mathrm{STE}}
=
\operatorname{sg}(\tilde{p}_i)
+
\left(p_i^{\tau}-\operatorname{sg}(p_i^{\tau})\right),
\label{eq:ste_weight}
\end{equation}
where $p_i^{\tau}=\mathrm{softmax}(\mathbf{z}/\tau)_i$ is a scaled routing probability with temperature $\tau$, and $\operatorname{sg}(\cdot)$ is the stop-gradient operator. In the forward pass, $\hat{p}_i^{\mathrm{STE}}$ reduces exactly to the standard sparse routing weight $\tilde{p}_i$. In the backward pass, the gradient of any loss $\mathcal{L}$ with respect to logit $z_j$ is:
\begin{equation}
\frac{\partial \mathcal{L}}{\partial z_j}
=
\sum_{i\in S}
\frac{\partial \mathcal{L}}{\partial \hat{p}_i^{\mathrm{STE}}}
\cdot
\frac{\partial p_i^{\tau}}{\partial z_j}.
\label{eq:ste_grad}
\end{equation}

Through STE, the router receives consistent data-driven feedback throughout training.

\subsection{Vision Encoder}
Intern-S1-Pro employs a Native Vision Transformer (ViT) as the vision encoder. The encoder processes images at native resolution, where the visual token count depends on the original input resolution rather than a fixed image size.
Such a design allows flexible handling of images with different spatial resolutions and preserves fine-grained spatial information in high-resolution inputs.
Visual tokens extracted from the ViT pass through a multilayer perceptron (MLP) projector that maps visual features into the embedding space of the language model, enabling joint multimodal reasoning.
The training of the encoder uses contrastive learning with large-scale image–text pairs.
Training data includes English caption datasets CC12M \cite{changpinyo2021cc12m}, LAION-COCO \cite{schuhmann2022laion5bopenlargescaledataset}, and SBU Caption \cite{ordonez2011im2text}, together with Chinese caption datasets LAION-2B-Multi \cite{relaion} and Wukong \cite{gu2022wukong100millionlargescale}. The combined corpus contains approximately 300 million image–text pairs. Such contrastive training improves visual representation quality and strengthens alignment between visual tokens and textual embeddings for downstream multimodal tasks.

\subsection{FoPE}
%\section{Bridging the Gap Between Discrete Tokens and Physical Reality}

Large Models have achieved remarkable success by processing information through discrete "tokens" — whether they represent text subwords, image patches, or audio frames. This tokenization paradigm inherently imposes a particle-like representation on all modalities, treating information as localized, discrete units. However, the physical world operates under fundamentally different principles: light exhibits wave-particle duality, sound propagates as continuous waveforms, and electromagnetic signals possess distinct spectral characteristics.
Traditional positional encoding methods, such as sinusoidal encodings or Rotary Position Embedding (RoPE) \cite{su2024roformer}, primarily serve to inject sequential order information into the model. While effective for capturing relative positions in text, these approaches are still weak in explicitly modeling the continuous, wave-like nature inherent in physical signals or the spectral properties that characterize multimodal data. This limitation creates a representational gap: language models process physical signals (images, audio, video) by flattening them into token sequences, thereby losing the rich spectral and wave-interference patterns that define their underlying physics.

Fourier Position Encoding (FoPE) \cite{hua2024fourier} addresses this fundamental limitation by reimagining how transformer models encode position and structure. Rather than treating positional information as merely an ordering mechanism, FoPE leverages the mathematical foundations of Fourier analysis to simultaneously capture both the discrete particle nature of tokens and the continuous wave characteristics of their interactions.

\begin{figure*}[ht]
    \centering
    \includegraphics[width=\textwidth]{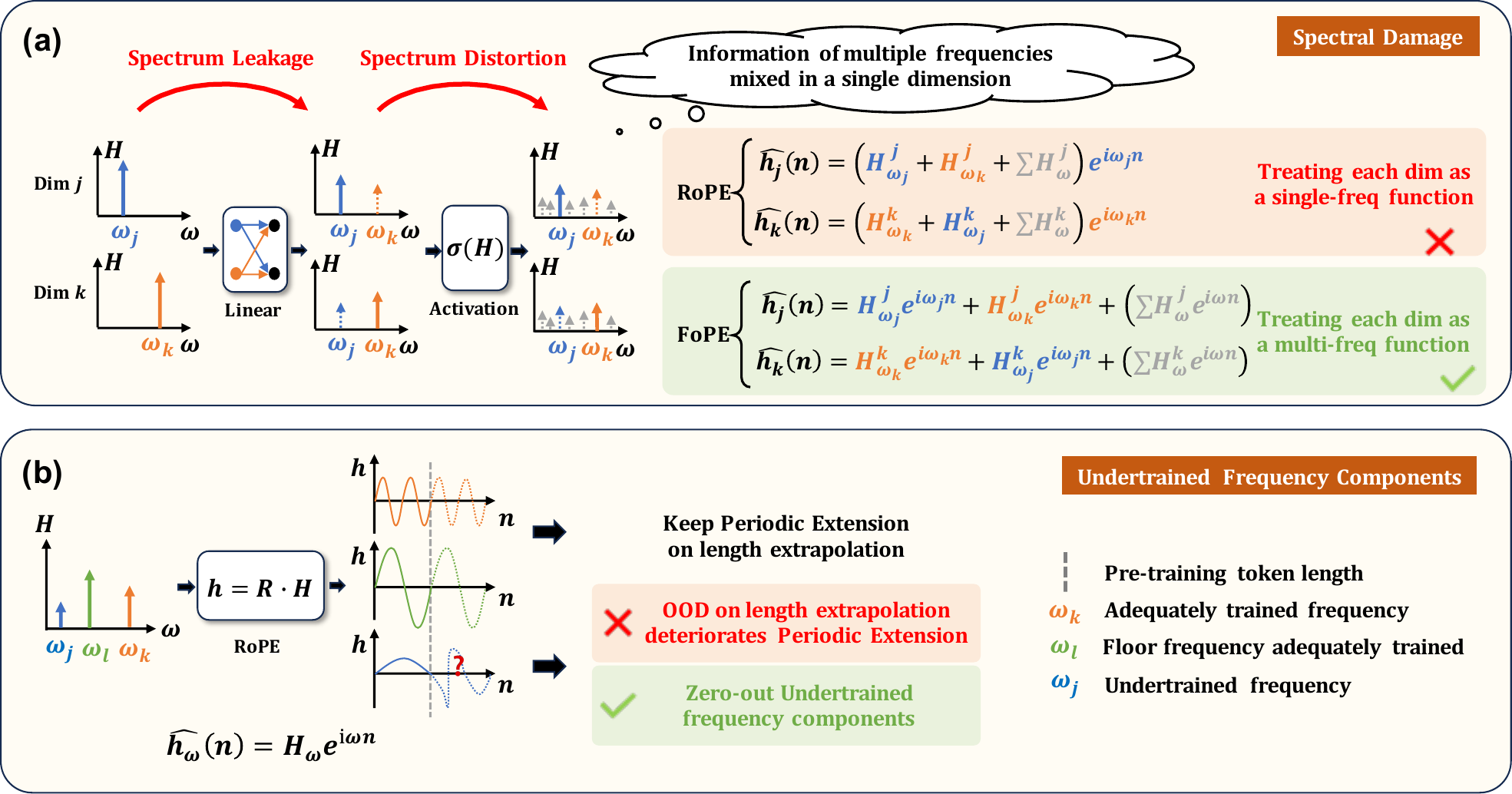}
    \caption{FoPE models each dimension as a Fourier series of different frequency components, thereby separating information more effectively and mitigating spectral damage. Inadequately trained frequency components are also clipped as their harmful influence on spectral.}
    \label{fig:framework}
\end{figure*}

\subsection{Time-series Encoder}

%\section{Time Series Module for Unified Scientific Time Series Modelling}

Time series is a core scientific data modality, capturing the temporal evolution of complex processes. Their extreme variability in rate, length, value, and dimensionality makes unified modelling challenging. Direct serialization into text tokens or conversion into images typically introduces information loss and limits numerical fidelity. The Intern-S1 family of scientific multimodal LLMs introduce a dedicated temporal modelling module that enables native time series understanding while preserving the reasoning and generalization strengths of LLMs.
% Time series constitute a fundamental data modality in scientific research, capturing the temporal evolution of complex physical and natural processes. Scientific time series signals often exhibit extreme variability in sampling rate, sequence length, numerical value, and dimensionality. This heterogeneity poses a significant challenge for unified modelling within a general foundation model framework. Direct serialization into text tokens or conversion into images typically introduces information loss and limits numerical fidelity, while traditional time series models are often task-specific and their effectiveness on non-periodic, heterogeneous scientific signals remains unclear. 

The time series module of Intern-S1-Pro, an enhanced successor to Intern-S1, expands both its disciplinary coverage and task diversity. Building upon its original support for astronomy, geoscience, and neuroscience applications, the enhanced module now incorporates additional domains such as physiological signal analysis and bioacoustics. This expansion enables a broader range of real-world scenarios, including electroencephalography-based depression detection, marmoset vocalization recognition, and electrocardiography abnormality monitoring. 

\begin{figure}[h]
    \centering
    \begin{subfigure}[t]{0.35\linewidth}
        \centering
        \includegraphics[width=\linewidth]{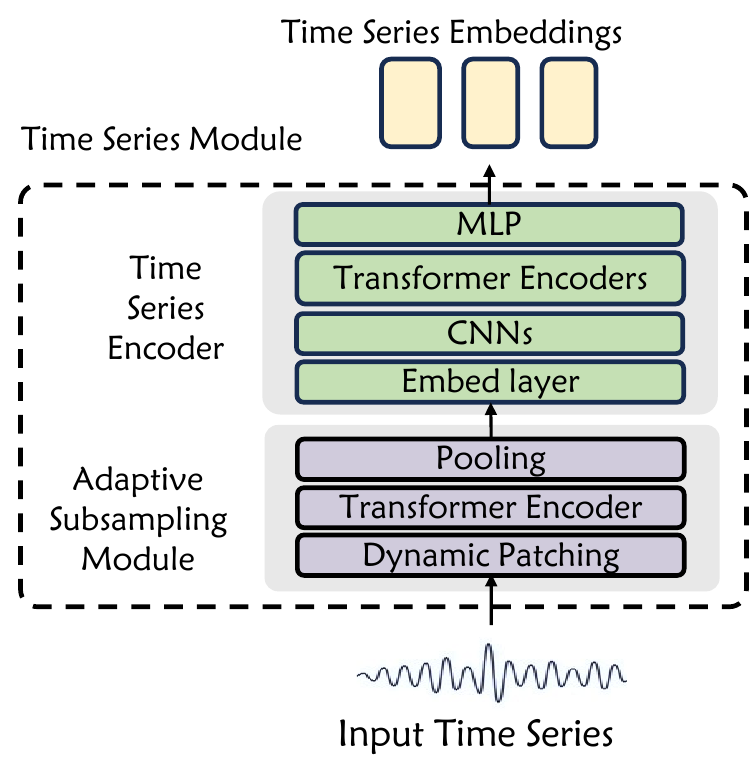}
        \caption{Structure of the time series module}
        \label{fig:overall}
    \end{subfigure}
    \hfill
    \begin{subfigure}[t]{0.6\linewidth}
        \centering
        \includegraphics[width=\linewidth]{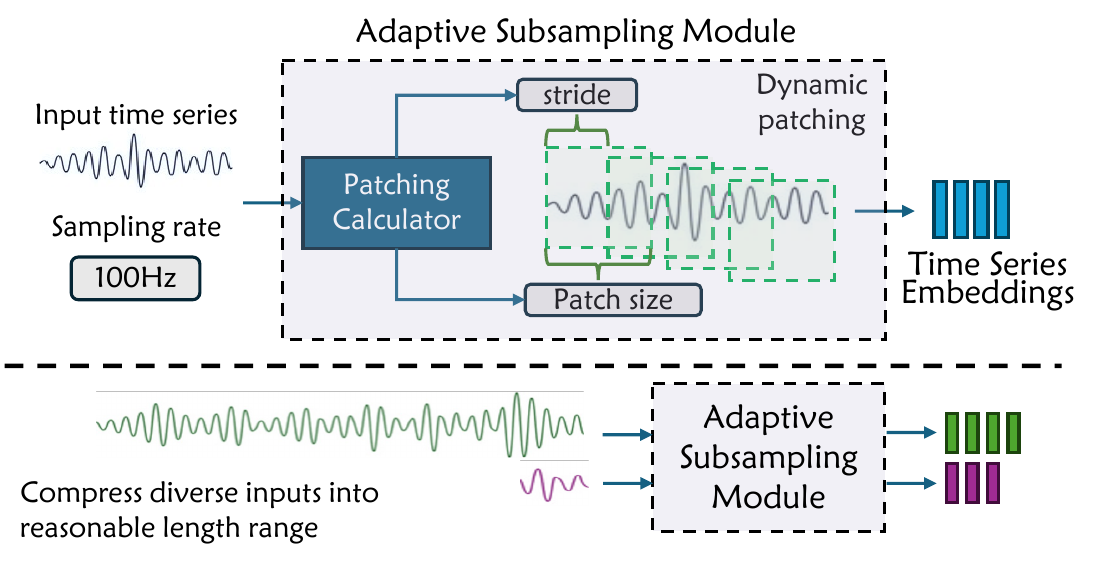}
        \caption{Illustration of the dynamic subsampling process}
        \label{fig:subsampling}
    \end{subfigure}
    \caption{Architecture and subsampling mechanism of the time series module.}
    \label{fig:ts_module}
\end{figure}

Furthermore, the time series module of Intern-S1-Pro features an upgraded architecture. As illustrated in Figure~\ref{fig:overall}, it consists of a novel adaptive subsampling module and a time series encoder.
% The core design of the time series module follows a hierarchical patch-based representation strategy. 
Given a continuous signal, the module first partitions it into local segments (patches), then captures local dynamics within each patch, and finally models long-range dependencies across segments. Rather than using pre-defined patch size and stride, these are adaptively determined based on the signal and its sampling rate, so that the number of temporal frames are kept within a controllable range (Figure~\ref{fig:subsampling}). The adaptive downsampling normalizes heterogeneous time series into a uniform representation space, enabling the encoder to handle sequences from $10^0$ to $10^6$ time steps while preserving structural features and computational efficiency.

The evaluation of model's time series capabilities will be detailed in the evaluation section.

% Given a continuous signal, the module first partitions it into a sequence of local segments (patches), capturing fine-grained dynamics within each segment before modelling long-range dependencies across segments. It's worth noting that . Instead, they are adaptively determined according to the input signal and its sampling rate, such that the number of resulting temporal frames remains within a controllable range, as illustrated in Figure~\ref{fig:subsampling}. This adaptive mechanism functions as a dynamic downsampling process that normalizes heterogeneous time series into a relatively uniform representation space. 
% As a result, the encoder is able to accommodate input sequences ranging from $10^0$ to $10^6$ time steps within a unified encoder while retaining their essential structural characteristics and maintaining computational efficiency.

%% file: sections/3.pretrain.tex
\section{Pre-training}

\begin{figure}[t]
    \centering
    \includegraphics[width=0.9\linewidth]{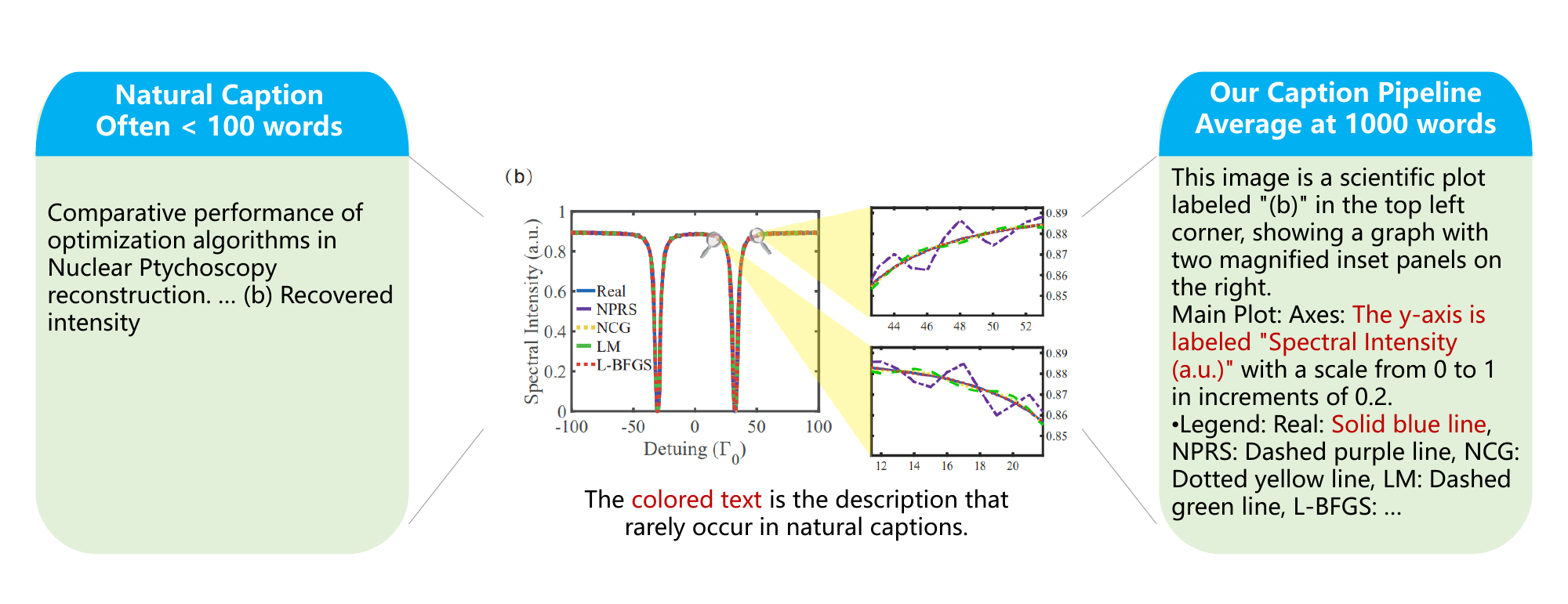}
    \caption{The comparison of natural caption (often occurs in scientific literature) and the desired dense caption for training VLM. The key is the text should explicitly refer the visual elements.}
    \label{fig:caption_case}
\end{figure}

\begin{figure}
    \centering
    \includegraphics[width=0.75\linewidth]{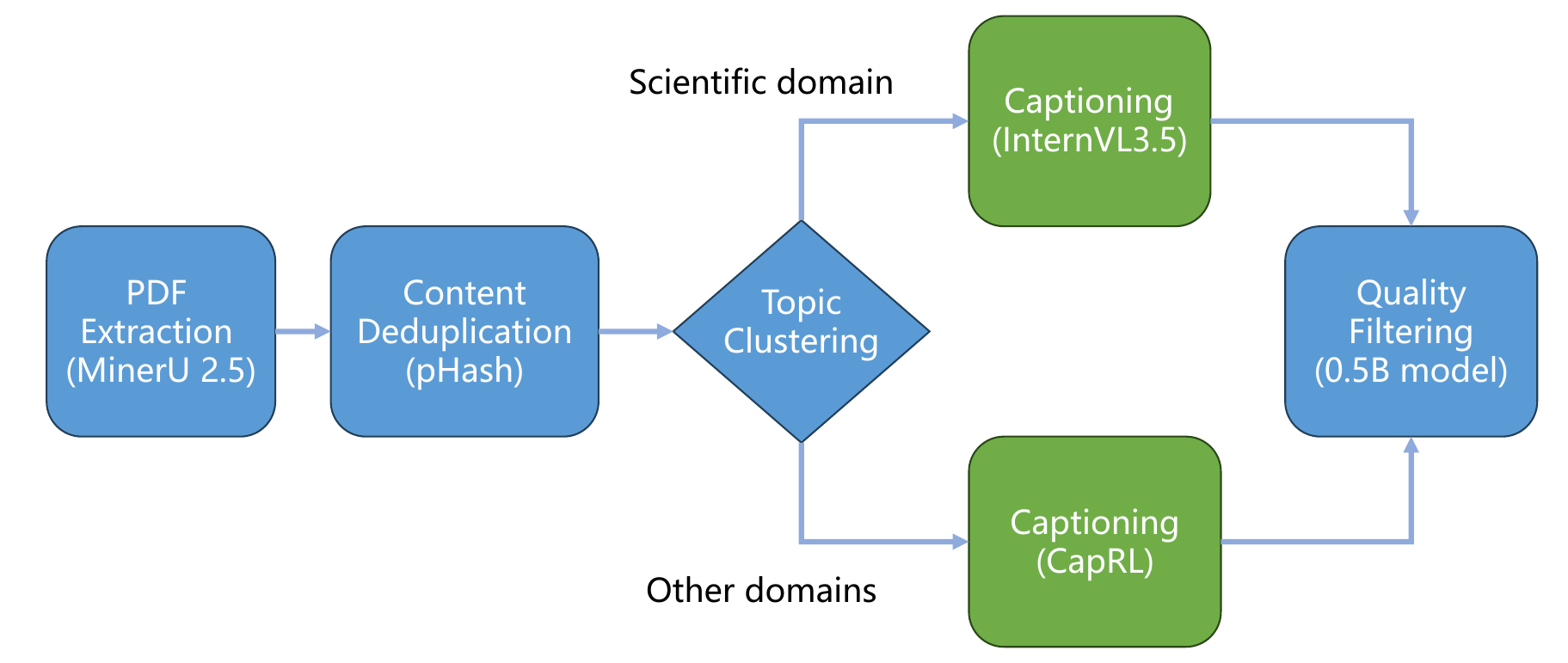}
    \caption{The workflow of caption pipeline used in data preparation, illustrating how high-quality aligned scientific multimodal data is produced in a efficient way.}
    \label{fig:placeholder}
\end{figure}

Intern-S1-Pro employs a total of 6T tokens of image-text and text data for continued pre-training. Compared to Intern-S1, a key upgrade lies in the caption data tailored for scientific images. As illustrated in Figure \ref{fig:placeholder}, the distribution of scientific images differs significantly from that of natural images, demanding higher accuracy in content understanding and greater attention to detail. Although high-quality images are available in public resources, acquiring high-quality image-text pairs is challenging. As shown in the figure, original captions in literature are often brief and lack alignment, their text is an extension of the image content rather than description. To address this, we have designed a dedicated caption pipeline to generate high-quality image-text pairs, thereby enhancing Intern-S1-Pro's understanding of scientific visual content.

\subsection{Caption Pipeline}

In the training of Vision-Language Models, high-quality image–text caption data serves as the core supervisory signal for cross-modal alignment. Existing open-source web caption datasets are largely derived from alt-text or surrounding webpage context\cite{schuhmann2022laion,kakaobrain2022coyo-700m}, which often exhibit limited image–text alignment and substantial semantic noise. Moreover, scientific images from web sources are insufficient in both scale and domain density, and we show an example in Figure~\ref{fig:caption_case}.

In contrast, PDFs represent the primary carrier of scientific visual content. They contain a wide range of high–information-density figures, including experimental results, statistical plots, structural diagrams, and formula derivations. As such, PDFs constitute a natural source of high-quality cross-modal alignment data, offering more systematic and domain-dense professional visual content. Therefore, beyond leveraging open-source web image–text caption datasets, we independently constructed a large-scale PDF data production pipeline tailored for scientific VLM training. This pipeline extracts sub-figures from massive PDF corpora and generates high-quality captions, enabling the systematic construction of dense, strongly aligned scientific image–text training data.

Specifically, we employ MinerU2.5 \cite{niu2025mineru2} for layout analysis and structural recognition, detecting and localizing figures, formulas, and tables, which are then cropped into standardized sub-image samples. We perform precise deduplication using perceptual hashing (pHash) to eliminate redundant visual content at scale. To further improve caption quality, we design a topic classification and model routing mechanism: scientific sub-images are described using InternVL3.5-241B to generate professional, domain-specific captions, while non-scientific sub-images are processed by CapRL-32B \cite{xing2025caprl}. CapRL (Captioning Reinforcement Learning) is a training framework that uses Reinforcement Learning with Verifiable Rewards (RLVR) to stimulate dense image caption capabilities. Based on this approach, we trained a CapRL model based on Qwen 2.5 VL 32B to generate high-quality, dense captions for general-purpose data across diverse domains. We incorporated this rich caption data into our pre-training pipeline, enhancing our model's ability to understand and describe visual content.

To enhance linguistic diversity, we adopt a multi-template randomized prompting strategy. Additionally, we introduce a 0.5B-parameter text quality discriminator to filter out garbled text, repetitive expressions, and low-information-density content, ensuring both high knowledge density and strong language quality in the final dataset.
This pipeline has been deployed at large scale on PDF corpora across life sciences, chemistry, earth sciences, and materials science, resulting in approximately 270B tokens of high-quality scientific image–text caption data. The resulting dataset provides abundant and high-quality supervisory signals for improving scientific understanding and reasoning capabilities in large-scale VLMs.

\subsection{Resolving conflicts between the scientific and textual data}

The integration of scientific data (such as experimental observations, chemical formulas, and structured literature) with general data (such as news and social media) presents significant challenges. Scientific data typically exhibits high logical determinism and structured features, while general data focuses on semantic depth and linguistic diversity~\cite{DBLP:journals/corr/abs-2211-09085}. Directly mixing these two types of data can lead to "distribution shift" and "negative transfer," resulting in logical confusion during model inference. To address this issue, Intern-S1-Pro adopts the following three major technical strategies:

\paragraph{Structured Scientific Data Transformation} Scientific data is typically represented in highly structured formats, which differ significantly from general data. To handle highly structured tabular information from databases like PubChem~\cite{DBLP:journals/nar/KimCCGHHLSTYZZB19}, we move beyond simple linearization and instead employ two methods: Template Construction and Task Form Transformation. Through template construction, heterogeneous input-output pairs are converted into grammatically correct, narrative text, ensuring that scientific data aligns with the representation style of general data. This transformation guarantees semantic consistency and minimizes information loss. Additionally, for abstract outputs like lists and matrices, which lack intuitive semantics, we combine domain-specific scientific priors to map numerical symbols to descriptive answers with actual scientific meaning. This allows the model to overcome symbolic barriers and better understand the underlying scientific logic and principles in the data.

\paragraph{Scientific Data Diversification} During the model training, data diversity is key to preventing overfitting. To enhance the diversity of training data, we upgrade both the input and output. First, to address the overfitting risk due to the high repetition of scientific data (e.g., similar protein sequences), we implement Prompt Diversification. By maintaining the core scientific knowledge, we provide dozens of varied instructions for the same concept, expanding the model's generalization boundaries. Second, to tackle the problem of overly simplistic outputs in scientific tasks (such as results containing only numerical values or conclusions), we introduce the Rollout mechanism. By combining scientific prior knowledge and leveraging a strong base model to assist in generating complete reasoning chains, we transform mere knowledge recall into logical deduction. This significantly enhances the model’s zero-shot reasoning ability in complex scientific scenarios.

\paragraph{System Prompt Isolation} Despite the data transformation and upgrade strategies, the differences between scientific and general data may still lead to conflicts during the training phase. To mitigate these conflicts and reduce negative impacts, we introduce the System Prompt Isolation strategy. During the training cycle, we inject mutually exclusive system-level prefixes for scientific and general data, creating independent contextual processing environments for the model. This strategy effectively reduces data conflicts, improves model stability, and enhances training effectiveness.

The three upgrade strategies outlined above effectively mitigate conflicts between scientific and general data. These strategies ensure more stable performance in multimodal training, enhancing the model's ability to perform across a wide range of tasks and reasoning scenarios.

%% file: sections/4.posttrain.tex
\section{Post-Training}

\subsection{Stable Mixed-Precision Reinforcement Learning for Sparse MoE Models}

Scaling reinforcement learning to trillion-parameter Mixture-of-Experts models presents formidable memory challenges. With Intern-S1-Pro featuring 4× the expert count of Intern-S1 while maintaining comparable activated parameters, the sheer volume of expert-layer parameters and activations creates substantial memory pressure even under expert parallelism. Following Intern-S1's approach, we adopt FP8 quantization for the RL phase, but the extreme sparsity of Intern-S1-Pro demands careful handling to prevent performance degradation under low-precision training.

% TODO：补一下引用
Prior work \cite{yao2025offpolicy} has identified training-inference engine discrepancy as a primary source of RL training instability. Several approaches have been proposed to address this: IcePop \cite{team2025every} employs importance sampling and masks tokens exhibiting large train-inference distribution shifts; A recent work \cite{ma2025stabilizing} introduces rollout router replay to ensure expert selection consistency between training and inference engines; MiniMax-M1 \cite{chen2025minimax} advocates for FP32 precision in the language modeling head to improve log-probability numerical accuracy; and KIMI-K2-Thinking \cite{moonshotai2025kimi} incorporates quantization-aware training (QAT) to adapt to low-precision representations. Building upon these foundations, we implement a comprehensive stabilization framework addressing operator-level precision discrepancies, keeping expert routing consistency, avoiding weight quantization mismatch, and applying importance sampling to loss functions.

First, we performed a systematic operator-by-operator comparison between the LMDeploy rollout engine and the XTuner training engine. We identified several numerically sensitive components that contributed disproportionately to divergence, including RMSNorm, router softmax, and positional embedding application. We then reduced the error between the two stacks by minimizing precision gaps in these kernels, ensuring that the rollout distribution is faithfully reflected during training. Second, to enforce expert consistency between rollout and training, we implement rollout router replay: for each token we record the selected expert indices per layer during rollout and replay the same routing decisions during policy updates. To avoid turning expert indices transfer into a bandwidth and latency bottleneck, we transmit these routing traces via Ray object references rather than sending them through the same HTTP channel used for response tokens. Third, we adopt a targeted mixed-precision scheme tailored to highly sparse MoE models. We observe that expert MLP layers account for the largest memory footprint, yet their GEMM operations are comparatively tolerant to reduced precision in our setting. We therefore quantize only expert linear layers to FP8, keep non-expert components in BF16, and use an FP32 LM head to improve the numerical fidelity of log-probabilities, following the general principle that small errors in log-probability estimation can be amplified by policy-gradient updates. This design preserves most of the memory and throughput benefits of FP8 while avoiding unnecessary degradation in sensitive parts of the computation graph. Finally, following IcePop, we modify the REINFORCE objective with dual importance sampling ratios. The loss is defined as

\begin{equation}
\mathcal{L}(\theta)
=
-
\mathbb{E}_{x \sim \mathcal{D},\, \{y_i\}_{i=1}^{G} \sim \pi_{\theta_{\text{rollout}}}(\cdot \mid x)}
\left[
\frac{1}{G}
\sum_{i=1}^{G}
\frac{1}{|y_i|}
\sum_{t=1}^{|y_i|}
\operatorname{sg}\!\left(\mathcal{M}(\rho_{i,t}; \alpha, \beta)\, r_{i,t}\right)
\cdot
\hat{A}_{i,t}
\cdot
\log \pi_{\theta}\!\left(y_{i,t}\mid x, y_{i,<t}\right)
\right].
\end{equation}

We define the first importance sampling ratio as $\rho_{i,t} = \frac{\pi_{\theta_{\text{train}}}(y_{i,t}\mid x, y_{i,<t})}{\pi_{\theta_{\text{rollout}}}(y_{i,t}\mid x, y_{i,<t})}$, calibrates for training-inference distribution mismatch. The second ratio $r_{i,t} =\frac{\pi_{\theta_{\text{new}}}(y_{i,t}\mid x, y_{i,<t})}{\pi_{\theta_{\text{old}}}(y_{i,t}\mid x, y_{i,<t})}.$, corrects for off-policy bias introduced by mini-batch updates during training. The masking function $\mathcal{M}(\rho_{i,t}; \alpha, \beta)$ is given by

\begin{equation}
\mathcal{M}(\rho_{i,t};\alpha,\beta)
=
\begin{cases}
\rho_{i,t}, & \alpha < \rho_{i,t} < \beta \\
0, & \text{otherwise}
\end{cases}
\end{equation}

\begin{figure*}[h]
    \centering
    \begin{subfigure}[t]{0.48\textwidth}
        \centering
        \includegraphics[width=\linewidth]{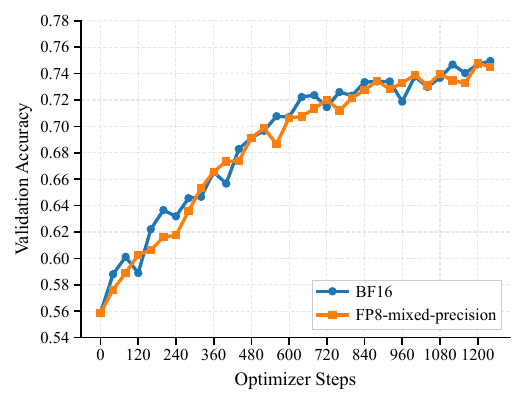}
        \caption{Validation accuracy across optimizer steps.}
        \label{fig:fp8_bf16_rl}
    \end{subfigure}
    \hfill
    \begin{subfigure}[t]{0.48\textwidth}
        \centering
        \includegraphics[width=\linewidth]{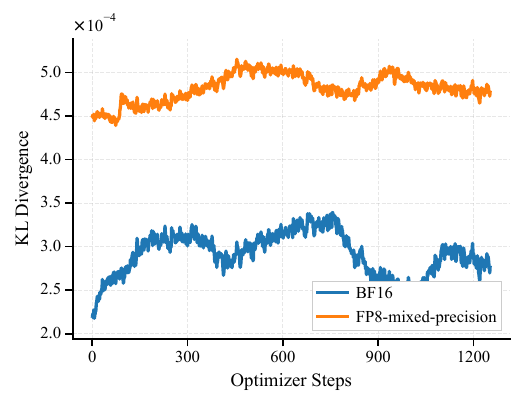}
        \caption{Log-prob KL curve between train engine and rollout engine.}
        \label{fig:mismatch_kl}
    \end{subfigure}
    \caption{Comparison between FP8 mixed-precision RL training and BF16 training on a 30B MoE model. The curves show the average accuracy across multiple validation benchmarks during RL optimization. FP8 training closely matches the BF16 baseline throughout training.}
    \label{fig:training_dynamics}
\end{figure*}

It is introduced for suppressing tokens whose training–rollout discrepancy is excessively large. For advantage estimation, we use a leave-one-out (LOO) baseline across the $G$ sampled responses:

\begin{equation}
\hat{A}_{i,t}
=
R_i - b_i,
\qquad
b_i
=
\frac{1}{G-1}\sum_{j\neq i} R_j,
\end{equation}
where $R_i$ denotes the sequence-level reward of sample $y_i$ and the same $\hat{A}_{i,t}$ is applied to all tokens $t$ within $y_i$. $\operatorname{sg}(\cdot)$ denotes the stop-gradient operator. 
With these changes, FP8 mixed-precision RL matches BF16 training behavior in practice as illustrated in Figure \ref{fig:training_dynamics}.

% \subsection{Async RL}
% TODO: 加上关于异步训练稳定性和效率的段落

%% file: sections/5.evaluation.tex
\newcommand{\open}[1]{\underline{#1}}
\newcommand{\best}[1]{\textbf{#1}}

\setlength{\aboverulesep}{0pt}
\setlength{\belowrulesep}{0pt}
\renewcommand{\arraystretch}{1.25} % 稍微增加行高

% --- 定义新的列格式 (去掉了 Task 列，Intern 变成了第2列) ---
% 1. Benchmark Name (左对齐)
% 2. InternS1 (蓝色背景, 居中)
% 3-6. 其他模型 (居中)
\definecolor{internblue}{RGB}{220,235,247}
\newcolumntype{L}[1]{>{\raggedright\arraybackslash}m{#1}}
\newcolumntype{C}[1]{>{\centering\arraybackslash}m{#1}}
\newcolumntype{B}[1]{>{\columncolor{internblue}\centering\arraybackslash}m{#1}}

% --- 定义子表标题样式 ---
\newcommand{\SubTableTitle}[1]{%
    \multicolumn{6}{l}{\hspace{-2pt}\textbf{#1}} \vspace{4pt} \\
}   

\section{Evaluation}

We conduct extensive experiments to evaluate Intern-S1-Pro across a wide range of benchmarks from two perspectives: scientific tasks and general-purpose tasks, covering both text-only and multimodal settings. In this section, we first introduce the evaluation setup, followed by a brief description of the benchmarks employed. We then compare the performance of Intern-S1-Pro with other state-of-the-art models.

\subsection{Evaluation Configuration}
We evaluate Intern-S1-Pro using three evaluation toolkits: OpenCompass\cite{2023opencompass}, VLMEvalKit\cite{duan2024vlmevalkit}, and AgentCompass\footnote{AgentCompass is an agent evaluation framework we developed, which will be released in the future.}. The evaluation setup is divided into two configurations: thinking and non-thinking. The specific configuration used for each benchmark is described in detail within the corresponding benchmark descriptions.

\begin{table}[h!]
\centering
\caption{Intern-S1-Pro Evaluation Configurations.}
\label{tab: eval-cfg}
\resizebox{0.4\linewidth}{!}{
\begin{tabular}{lcc}
\toprule
 & \textbf{Thinking} &  \textbf{Non-Thinking} \\
\midrule
max tokens & 65536  & 32768 \\
temperature & 0.8 & 0 \\
top\_p & 0.95 &1.0 \\
top\_k &50 &1 \\
\bottomrule
\end{tabular}
}
\end{table}

\begin{table*}[t]
\centering
\caption{Comprehensive performance comparison across scientific and general benchmarks. The highest scores are highlighted in \best{bold}, and the second-highest scores are \open{underlined}.}
\label{tab:main_results}
\small
\setlength{\tabcolsep}{4pt} % 稍微减小列间距以防太宽

\begin{adjustbox}{max width=\textwidth}
\begin{tabular}{
  l l
  B{2.2cm}
  C{2.8cm}
  C{2.2cm}
  C{2.2cm}
  C{2.2cm}
}

% ================= SUB-TABLE 2: Scientific Task =================
\multicolumn{6}{c}{}\\[-0.5em]
\SubTableTitle{Scientific Tasks}
\toprule
  \multirow{2}{*}{\textbf{Benchmark}} & \multirow{2}{*}{\textbf{Description}} &
  \textbf{Intern-S1-Pro} &
  \textbf{Qwen3-VL-235B-Thinking} &
  \textbf{Kimi-K2.5} &
  \textbf{GPT-5.2} &
  \textbf{Gemini-3-Pro}\\
  & &
  1T-A22B & 235B-A22B & 1T-A32B & --  & -- \\
\midrule
  SciReasoner         & Scientific Reasoning & \open{\best{55.5}} & 11.9 & 15.3 & 13.6 & 14.7 \\
  SFE                 & Scientific Multimodal Tasks & 52.7 & 41.4 & \open{53.7} & 47.5 & \best{58.9}  \\
  SmolInstruct        & Small Molecule & \open{\best{74.8}} & 36.6 & 53.5 & 48.2 & 58.3  \\
  MatBench            & Materials Property Prediction & \open{\best{72.8}} & 49.7 & 60.0 & 53.6 & 64.9 \\
  Mol-Instructions    & Bio-molecular Instruction & \open{\best{48.8}} & 8.9 & 20.0 & 12.3 & 34.6 \\
  MicroVQA            & Biological Microscopy & \open{63.3} & 53.8 & 55.4 & 60.4 & \best{69.0}  \\
  Biology-Instruction & Multi-Omics Sequence & \open{\best{52.5}} & 6.2 & 10.7 & 10.2 & 12.0 \\
  XLRS-Bench          & Remote Sensing & \open{\best{52.8}} & 51.2 & 46.4 & 50.4 & 51.8  \\
  MSEarth-MCQ         & Earth Science & \open{65.2} & 52.7 & 61.9 & 62.6 & \best{65.8} \\
\bottomrule

% ================= SUB-TABLE 1: General Task =================
\multicolumn{6}{c}{}\\[-0.5em]
\SubTableTitle{General Tasks}
\toprule
  \multirow{2}{*}{\textbf{Benchmark}} & \multirow{2}{*}{\textbf{Description}} &
  \textbf{Intern-S1-Pro} &
  \textbf{Qwen3-VL-235B-Thinking} &
  \textbf{Kimi-K2.5} &
  \textbf{GPT-5.2} &
  \textbf{Gemini-3-Pro}\\
  & &
  1T-A22B & 235B-A22B & 1T-A32B & --  & -- \\
\midrule
  MMMU-Pro & Knowledge \& Reasoning & 72.8 & 69.9 & \open{78.5} & 79.5 & \best{81.0} \\
  MMLU-Pro & Knowledge \& Reasoning & 86.6 & 83.4 & \open{87.1} & 85.9 & \best{89.3}  \\
  % GPQA & Knowledge \& Reasoning & 80.2 & 78.4 & \open{87.6} & \best{92.4} & 91.5 \\
  % MathVista-MINI & Math Reasoning & 84.7 & 84.8 & \open{\best{89.7}} & 82.6 & 89.4  \\
  AIME-2025 & Math Reasoning & 93.1 & 90.0 & \open{96.1} & \best{100.0} & 95.0 \\
  IMO-Answer-Bench & Math Reasoning & 77.3 & 72.3 & \open{81.8} & \best{86.3} & 81.3  \\
  % IPhO-2025 & Physics Reasoning & 21.1 & 17.2 & \open{22.0} & 21.8 & \best{23.5} \\
  RefCOCO-avg & Visual Grounding & \open{\best{91.9}} & 91.1 & 87.8 & 54.9 & 76.2  \\
  IFBench & Instruction Following & \open{71.2} & 58.7 & 69.7 & \best{75.4} & 70.4  \\
  OCRBench V2 (ENG / CHN) & OCR & 60.1 / 60.6 & \open{66.8} / \open{\best{63.8}} & 64.2 / 57.4 & 56.4 / 54.6 & \best{68.0} / 52.5 \\
  SArena (Icon) & SVG Generation & \open{\best{83.5}} & 76.3 & 77.3 & 80.5 & 82.6  \\
  LCB V6 & Code & 74.3 & 72.0 & \open{85.0} & \best{87.7} & 86.9 \\
  GAIA (Text-Only) & Agent & 77.4 & 47.8 & \open{\best{79.9}} & 71.1 & 75.5 \\
  $\tau^2$\text{-Bench} & Agent & \open{80.9} & 57.4 & 76.8 & 76.6 & \best{85.4}  \\
  ScreenSpot V2 & Agent \& Grounding  & \open{93.6} & 92.8 & 92.4 & 49.4 & \best{94.7}  \\
\bottomrule

\end{tabular}
\end{adjustbox}
\end{table*}

\subsection{Benchmarks}

\subsubsection{Scientific benchmarks}

\textbf{SciReasoner} \cite{wang2025scireasonerlayingscientificreasoning} evaluates scientific reasoning across ten diverse disciplines, including physics, chemistry, and medicine, 9 domains in total and 149 concrete tasks. It consists of a unified suite of ten sub-benchmarks with varying question formats such as multiple-choice, fill-in-the-blank, and protocol-based procedural questions, designed to assess both knowledge retrieval and complex deductive reasoning.  We use the \textbf{non-thinking} configuration to evaluate the benchmark.

\textbf{SFE} \cite{zhou2025scientistsexamprobingcognitive} is an expert-level benchmark comprising 830 verified visual question answering (VQA) pairs across 66 multimodal tasks. Spanning five high-value scientific disciplines, the dataset utilizes authentic raw scientific data formats to probe the cognitive abilities of models in perception, understanding, and advanced reasoning. We use the \textbf{thinking} configuration to evaluate the benchmark.

\textbf{SmolInstruct} \cite{yu2024llasmoladvancinglargelanguage} is a large-scale chemistry-specific dataset that features 14 selected tasks and over three million samples for instruction tuning. The benchmark covers meaningful chemical applications, including forward synthesis and property prediction, providing a high-quality foundation for evaluating multi-step reasoning in molecular science. We use the \textbf{non-thinking} configuration to evaluate the benchmark.

\textbf{MatBench} \cite{Dunn_2020} provides a curated test suite of 13 machine learning tasks for materials property prediction, with dataset sizes ranging from 312 to 132,000 samples. Derived from 10 density functional theory and experimental sources, it standardizes the evaluation of model performance on diverse crystalline and molecular materials properties. We use the \textbf{non-thinking} configuration to evaluate the benchmark.

\textbf{Mol-Instructions} \cite{fang2024molinstructionslargescalebiomolecularinstruction} is designed to bridge the gap in specialized LLM training through three primary categories: molecule-oriented, protein-oriented, and biomolecular text-oriented tasks. It includes a vast collection of instruction-following pairs that facilitate the model's proficiency in handling complex biomolecular structures and functional descriptions. We use the \textbf{non-thinking} configuration to evaluate the benchmark.

\textbf{MicroVQA} \cite{burgess2025microvqamultimodalreasoningbenchmark} focused on microscopy-based research, consists of 1,042 expert-curated multiple-choice questions across diverse imaging modalities. The benchmark assesses three critical reasoning capabilities within biological workflows: expert image understanding, hypothesis generation, and experimental proposal. We use the \textbf{non-thinking} configuration to evaluate the benchmark.

\textbf{Biology-Instruction} \cite{he2025biologyinstructionsdatasetbenchmarkmultiomics} is a multi-omics benchmark that evaluates the sequence understanding capabilities of models across diverse biological scales. It integrates biological sequence-based prediction tasks with advanced reasoning requirements, challenging models to interpret complex genomic, transcriptomic, and proteomic data. We use the \textbf{non-thinking} configuration to evaluate the benchmark.

\textbf{XLRS-Bench} \cite{Wang_2025_CVPR} focused on extremely large, ultra-high-resolution remote sensing (RS) imagery; this benchmark defines 16 sub-tasks to evaluate 6 types of perceptual and 4 types of reasoning abilities. It challenges MLLMs to process complex semantic relationships and facilitate real-world decision-making in high-resolution geospatial scenarios. We use the \textbf{thinking} configuration to evaluate the benchmark.

\textbf{MSEarth-MCQ} \cite{zhao2025msearthmultimodalscientificdataset} is a multimodal scientific dataset curated from high-quality earth science publications to uncover natural phenomena. It transforms complex scientific figures and remote sensing data into multiple-choice questions, emphasizing the alignment between visual perception and domain-specific logical reasoning. We use the \textbf{non-thinking} configuration to evaluate the benchmark.

\subsubsection{General benchmarks}

\textbf{MMMU-Pro} \cite{yue2025mmmuprorobustmultidisciplinemultimodal} is a robust extension of the MMMU benchmark. MMMU-Pro introduces more challenging multidisciplinary multimodal tasks. It emphasizes expert-level understanding and complex reasoning across a wide array of professional domains, utilizing high-resolution images and specialized knowledge. We use the \textbf{thinking} configuration to evaluate the benchmark.

\textbf{MMLU-Pro} \cite{wang2024mmluprorobustchallengingmultitask} enhances the original MMLU by increasing the number of choices and focusing on more challenging, reasoning-intensive questions. It covers a broad range of subjects, requiring models to demonstrate deeper multi-task language understanding and more robust problem-solving skills. We use the \textbf{thinking} configuration to evaluate the benchmark.

\textbf{AIME-2025} comprises 30 high-level competition problems from the 2025 American Invitational Mathematics Examination. The tasks require olympiad-level mathematical reasoning, where models must provide precise integer answers (000-999) for complex problems in algebra, geometry, and number theory. We use the \textbf{thinking} configuration to evaluate the benchmark.

\textbf{IMO-Answer-Bench} \cite{luong2025robustmathematicalreasoning} is designed to evaluate robust mathematical reasoning. This benchmark features 400 diverse Olympiad-level problems with verifiable short answers. It prioritizes genuine problem-solving capabilities and aims to deter memorization by using challenging, competition-grade mathematical content. We use the \textbf{thinking} configuration to evaluate the benchmark.

\textbf{RefCOCO} \cite{kazemzadeh-etal-2014-referitgame} is a classic benchmark for referring expression comprehension. RefCOCO requires models to localize specific objects in natural scenes based on natural language descriptions. It tests the precise alignment between linguistic instructions and visual grounding in diverse photographic contexts. We use the \textbf{non-thinking} configuration to evaluate the benchmark.

\textbf{IFBench} \cite{pyatkin2025generalizingverifiableinstructionfollowing} evaluates the generalization of precise instruction following across 58 new and challenging verifiable out-of-domain constraints. The tasks include complex manipulations of text, formatting, and character-level constraints, testing the model's ability to adhere to strict output requirements. We use the \textbf{thinking} configuration to evaluate the benchmark.

\textbf{OCRBench V2} \cite{fu2025ocrbenchv2improvedbenchmark} assesses large multimodal models on visual text localization and reasoning. It features a diverse set of images requiring fine-grained OCR capabilities, including the recognition and spatial grounding of text in various complex real-world scenes. We use the \textbf{non-thinking} configuration to evaluate the benchmark.

\textbf{SArena} \cite{wang2026internsvgunifiedsvgtasks} is part of the InternSVG suite. SArena offers unified task definitions for Scalable Vector Graphics (SVG) understanding, editing, and generation. It spans a difficulty spectrum from simple icons to complex scientific diagrams and animations, providing standardized evaluation protocols for SVG-related multimodal tasks. We adopt the icon subset and use the \textbf{thinking} configuration to evaluate the benchmark.

\textbf{LCB V6} \cite{jain2024livecodebenchholisticcontaminationfree} is the latest version of LiveCodeBench, which provides a holistic and contamination-free evaluation for code generation by continuously collecting problems from competitive programming platforms. It assesses models on code completion, bug fixing, and execution reasoning across multiple programming languages. We use the \textbf{thinking} configuration to evaluate the benchmark.

\textbf{GAIA} \cite{mialon2023gaiabenchmarkgeneralai} evaluates models on real-world tasks that are conceptually simple for humans but challenging for AI. It requires the integration of tool use, multi-modal reasoning, and multi-step planning to solve practical problems across diverse digital environments. We use the \textbf{thinking} configuration to evaluate the benchmark. We implemented a simple agent workflow equipped with web search tools (Google \& Jina) to evaluate the GAIA benchmark.

\textbf{Tau2-Bench} \cite{barres2025tau2benchevaluatingconversationalagents} evaluates conversational agents in dual-control environments, simulating technical support scenarios like telecom services. It requires models to coordinate actions with active users and interact with shared world states to resolve complex service issues. We use the \textbf{thinking} configuration to evaluate the benchmark. The evaluation settings follow the official settings.

\textbf{ScreenSpot} V2 \cite{wu2024osatlasfoundationactionmodel} is an evaluation benchmark for GUI grounding, comprising over 1,200 instructions across iOS, Android, macOS, Windows, and Web environments. It assesses the model's ability to precisely locate UI elements, such as text, icons, and widgets, based on natural language commands. We use the \textbf{non-thinking} configuration to evaluate the benchmark.

\subsection{Main Results}

\textbf{Overall Performance.} As shown in Table~\ref{tab:main_results}, Intern-S1-Pro firmly demonstrates highly competitive capabilities with the first tier open-source models. Notably, its scientific reasoning capabilities are ahead of leading closed-source models. In the scientific tasks evaluation, Intern-S1-Pro significantly outperforms proprietary models like Gemini-3-Pro and GPT-5.2 on multiple benchmarks. For example, it achieves an outstanding score of 55.5 on SciReasoner, compared to 14.7 for Gemini-3-Pro and 13.6 for GPT-5.2. Furthermore, Intern-S1-Pro secures the top position on diverse scientific benchmarks including SmolInstruct (74.8), MatBench (72.8), Mol-Instructions (48.8), Biology-Instruction (52.5), and XLRS-Bench (52.8). In general tasks, Intern-S1-Pro maintains strong performance, achieving 93.1 on AIME-2025 and 86.6 on MMLU-Pro, matching or exceeding the capabilities of strong open-source models like Qwen3-VL-235B-Thinking.

\textbf{Improvements over Intern-S1.} Compared to the previous generation, Intern-S1, Intern-S1-Pro delivers substantial improvements in three main aspects. First, its general capabilities have been consistently improved; for instance, performance on AIME-2025 increased from 86.0 to 93.1, and MMLU-Pro improved from 83.5 to 86.6. Second, Intern-S1-Pro significantly increases its coverage of scientific tasks. Building upon the foundational scientific benchmarks evaluated in Intern-S1, the new model extends to more diverse and challenging domains such as SciReasoner, Mol-Instructions, and Biology-Instruction, while maintaining leading performance. Finally, Intern-S1-Pro introduces newly enhanced agent capabilities. It demonstrates robust multi-step planning and environmental grounding, achieving 77.4 on GAIA (Text-Only), 80.9 on $\tau^2$-Bench, and 93.6 on ScreenSpot V2, marking a significant step forward in practical agentic applications.

\begin{table}[h!]
\centering
\caption{Results on a subset of the SciTS benchmark. F1 scores are reported.}
\label{tab: res}
\resizebox{\linewidth}{!}{
\begin{tabular}{llccccccccc}
\toprule
 & \textbf{SciTS Task ID} &  \textbf{ASU01} & \textbf{ASU03} & \textbf{BIU01} & \textbf{BIU03} & \textbf{EAU01} & \textbf{MEU01} & \textbf{NEU06} & \textbf{PHU01} & \textbf{PHU04} \\
\midrule
\multirow{3}{*}{Text LLM} & GPT-4.1-mini        & 67.2 & 15.6 & 0.2  & 12.7 & 67.0 & 44.0 & 16.1 & 24.0 & 52.7 \\
                           & Gemini2.5-Flash     & 64.1 & 16.3 & 1.5  & 12.4 & 67.6 & 60.9 & 5.8  & 20.7 & 64.8 \\
                           & DeepSeek-V3         & 1.1  & 12.3 & 0.0  & 5.8  & 40.2 & 59.3 & 13.6 & 28.9 & 50.7 \\
\midrule
\multirow{2}{*}{VL LLM}   & GPT-5-mini          & 65.7 & 18.9 & 0.8  & 17.9 & 67.6 & 30.4 & 13.3 & 21.4 & 47.8 \\
                           & Gemini2.5-Flash     & 61.6 & 15.2 & 0.9  & 8.3  & 72.5 & 64.1 & 11.6 & 22.7 & 59.0 \\
                           \midrule
                           \rowcolor{internblue} & \textbf{Intern-S1-Pro}       & \textbf{98.0} & \textbf{75.9} & \textbf{20.8} & \textbf{88.3} & \textbf{99.5} & \textbf{65.6} & \textbf{71.3} & \textbf{36.8} & \textbf{93.2} \\
\bottomrule
\end{tabular}
}
\end{table}

\subsection{Time Series Results}

Table~\ref{tab: res} reports results on a subset of the SciTS benchmark~\citep{wu2025scits}, demonstrating the advantages of incorporating a dedicated time series processing module into multimodal LLMs. Compared to both Text LLMs (such as GPT-4.1-mini and DeepSeek-V3) and Vision-Language LLMs (such as GPT-5-mini and Gemini2.5-Flash), Intern-S1-Pro exhibits significantly superior performance across diverse scientific time series tasks. For instance, on the EAU01 task, Intern-S1-Pro achieves an F1 score of 99.5, vastly outperforming other models. Similarly, substantial improvements are observed on challenging tasks like BIU03 and PHU04, verifying the effectiveness of the proposed dynamic subsampling process and the dedicated time series encoder in capturing complex temporal dynamics.

\begin{table}[h!]
\centering
\caption{Comparison between specialized model (Biology-Instruction \cite{he2025biologyinstructionsdatasetbenchmarkmultiomics}) and Intern-S1-Pro on biological tasks.}
\label{tab:bio_eval}
\resizebox{0.7\linewidth}{!}{
\begin{tabular}{lcc}
\toprule
\textbf{Dataset} & \textbf{Biology-Instruction} & \cellcolor{internblue}\textbf{Intern-S1-Pro} \\
\midrule
DNA-cpd & 44.54 & \cellcolor{internblue}\textbf{54.60} \\
DNA-emp & 8.10 & \cellcolor{internblue}\textbf{14.02} \\
DNA-pd & 58.18 & \cellcolor{internblue}\textbf{82.65} \\
DNA-tf-h & 24.45 & \cellcolor{internblue}\textbf{54.11} \\
DNA-tf-m & 39.91 & \cellcolor{internblue}\textbf{60.80} \\
Multi\_sequence-antibody\_antigen & 10.26 & \cellcolor{internblue}\textbf{44.76} \\
Multi\_sequence-promoter\_enhancer\_interaction & \textbf{4.77} & \cellcolor{internblue}-1.30 \\
Multi\_sequence-rna\_protein\_interaction & \textbf{74.26} & \cellcolor{internblue}58.51 \\
DNA-enhancer\_activity & 53.28 & \cellcolor{internblue}\textbf{55.16} \\
RNA-CRISPROnTarget & 3.77 & \cellcolor{internblue}\textbf{15.69} \\
Protein-Fluorescence & 2.57 & \cellcolor{internblue}\textbf{78.14} \\
Protein-Stability & 60.25 & \cellcolor{internblue}\textbf{60.82} \\
Protein-Thermostability & 45.07 & \cellcolor{internblue}\textbf{59.56} \\
RNA-Isoform & 59.01 & \cellcolor{internblue}\textbf{82.95} \\
RNA-MeanRibosomeLoading & 47.64 & \cellcolor{internblue}\textbf{52.41} \\
RNA-ProgrammableRNASwitches & 26.65 & \cellcolor{internblue}\textbf{33.97} \\
RNA-Modification & \textbf{59.06} & \cellcolor{internblue}57.77 \\
Protein-Solubility & 63.02 & \cellcolor{internblue}\textbf{67.60} \\
RNA-NoncodingRNAFamily & \textbf{63.09} & \cellcolor{internblue}34.50 \\
Protein-FunctionEC & 19.79 & \cellcolor{internblue}\textbf{72.70} \\
Multi\_sequence-sirnaEfficiency & 56.31 & \cellcolor{internblue}\textbf{62.05} \\
\midrule
\textbf{AVG score} & 39.24 & \cellcolor{internblue}\textbf{52.45} \\
\bottomrule
\end{tabular}
}
\end{table}

\subsection{Specializable Generalist could be Better: A Case Study in Biology}
\label{sec:bio_case}

During the development of Intern-S1-Pro, we observed that a larger, more general foundation model has the potential to further enhance capabilities in highly specialized domains. We present a case study comparing Intern-S1-Pro against a specialized domain model, Biology-Instruction, on a series of biological sequence and structure tasks. As shown in Table~\ref{tab:bio_eval}, Intern-S1-Pro achieves significantly better comprehensive performance. 

Notably, both models were trained on the same underlying dataset, and we only upgraded the data for Intern-S1-Pro to feature more fluent text expression, while the core biological information remained identical. Benefiting from the enhanced intelligence brought by its larger model size and stronger general reasoning capabilities, Intern-S1-Pro manages to extract and utilize the same specialized data much more effectively. For instance, on the Protein-Fluorescence task, Intern-S1-Pro achieves a score of 78.14 compared to Biology-Instruction's 2.57, and on Protein-FunctionEC, it scores 72.70 compared to 19.79. These results strongly suggest that the integration of general and specialized capabilities is not merely an aggregation of functions, but a synergistic process that fundamentally promotes the intelligence and problem-solving capacity of the model in professional domains.

%% file: sections/6.conclusion.tex
\section{Conclusion}

In this report, we introduced Intern-S1-Pro, a trillion-parameter scientific multimodal foundation model designed to advance the frontiers of AI in scientific discovery. Building upon the strong foundation of Intern-S1, we scaled the model through a novel expert expansion strategy combined with Grouped Routing. This architectural innovation not only ensures efficient load balancing across devices but also significantly enhances training stability, mitigating the risks of expert homogenization and training instability often observed in large-scale MoE models.

To further bolster the model's scientific understanding, we conducted continued pre-training on 6T tokens of high-quality multimodal data. A critical component of this process was the development of a specialized caption pipeline tailored for scientific imagery. By generating precise, alignment-focused captions for scientific figures, we overcame the limitations of existing datasets and substantially improved the model's ability to interpret complex scientific visual content.

Our extensive evaluations demonstrate that Intern-S1-Pro achieves state-of-the-art performance across a wide range of scientific benchmarks, exhibiting robust reasoning capabilities and deep domain knowledge. These results underscore the effectiveness of our architectural and data strategies. Moving forward, we aim to further expand the model's capabilities into more specialized scientific domains for the acceleration of scientific discovery.

\subsubsection*{Author Contributions}
The authors are listed in alphabetical order by their last names.

Yicheng Zou, Dongsheng Zhu, Lin Zhu, Tong Zhu, Yunhua Zhou, Peiheng Zhou, Xinyu Zhou, Dongzhan Zhou, Zhiwang Zhou, Yuhao Zhou, Bowen Zhou, Zhanping Zhong, Zhijie Zhong, Haiteng Zhao, Penghao Zhao, Xiaomeng Zhao, Zhiyuan Zhao, Yechen Zhang, Jin Zhang, Wenwei Zhang, Hongjie Zhang, Zhuo  Zhang, Wenlong Zhang, Bo Zhang, Chao Zhang, Chen Zhang, Yuhang Zang, Fei Yuan, Jiakang Yuan, Jiashuo Yu, Jinhui Yin, Haochen Ye, Qian Yao, Bowen Yang, Danni Yang, Kaichen Yang, Ziang Yan, Jun Xu, Yicheng Xu, Wanghan Xu, Xuenan Xu, Chao Xu, Ruiliang Xu, Shuhao Xing, Long Xing, Xinchen Xie, Ling-I Wu, Zijian Wu, Zhenyu Wu, Lijun Wu, Yue Wu, Jianyu Wu, Wen Wu, Fan Wu, Xilin Wei, Qi Wei, Bingli Wang, Rui Wang, Ziyi Wang, Zun Wang, Yi Wang, Haomin Wang, Yizhou Wang, Lintao Wang, Yiheng Wang, Longjiang Wang, Bin Wang, Jian Tong, Zhongbo Tian, Huanze Tang, Chen Tang, Shixiang Tang, Yu Sun, Qiushi Sun, Xuerui Su, Qisheng Su, Chenlin Su, Demin Song, Jin Shi, Fukai Shang, Yuchen Ren, Pengli Ren, Xiaoye Qu, Yuan Qu, Jiantao Qiu, Yu Qiao, Biqing Qi, Runyu Peng, Tianshuo Peng, Jiahui Peng, Qizhi Pei, Zhuoshi Pan, Linke Ouyang, Wenchang Ning, Yichuan Ma, Zerun Ma, Ningsheng Ma, Runyuan Ma, Chengqi Lyu, Haijun Lv, Han Lv, Lindong Lu, Kuikun Liu, Jiangning Liu, Yuhong Liu, Kai Liu, Hongwei Liu, Zhoumianze Liu, Mengjie Liu, Ziyu Liu, Wenran Liu, Yang Liu, Liwei Liu, Kaiwen Liu, Junyao Lin, Junming Lin, Tianyang Lin, Dahua Lin, Jianze Liang, Linyang Li, Peiji Li, Zonglin Li, Zehao Li, Pengze Li, Guoyan Li, Lingkai Kong, Linglin Jing, Zhenjiang Jin, Feifei Jiang, Qian Jiang, Junhao Huang, Zixian Huang, Haian Huang, Zhouqi Hua, Ermo Hua, Han Hu, Linfeng Hou, Yinan He, Conghui He, Tianyao He, Xu Guo, Qipeng Guo, Aijia Guo, Yuzhe Gu, Lixin Gu, Jingyang Gong, Qiming Ge, Jiaye Ge, Songyang Gao, Jianfei Gao, Xinyu Fang, Caihua fan, Yue Fan, Yanhui Duan, Zichen Ding, Shengyuan Ding, Ning Ding, Xuanlang Dai, Erfei Cui, Ganqu Cui, Pei Chu, Tao Chu, Guangran Cheng, Yu Cheng, Kai Chen, Yongkang Chen, Chiyu Chen, Guanzhou Chen, Qiaosheng Chen, Sitao Chen, Xin Chen, Haojiong Chen, Yicheng Chen, Weihan Cao, Yuhang Cao, Qinglong Cao, Lei Bai